\documentclass[10pt,twocolumn,letterpaper]{article}

\usepackage{cvpr}              

\usepackage{microtype}

\renewcommand{\paragraph}[1]{\noindent\textbf{#1} }

\usepackage{booktabs}
\usepackage{xcolor}
\usepackage{colortbl}
\usepackage{multirow}
\usepackage{graphicx}
\usepackage{makecell}
\usepackage{verbatim}
\usepackage{textcomp}
\usepackage{float}

\usepackage{stfloats}

\definecolor{cvprblue}{rgb}{0.21,0.49,0.74}
\usepackage[pagebackref,breaklinks,colorlinks,allcolors=cvprblue]{hyperref}
\def\paperID{} 
\def\confName{CVPR}
\def\confYear{2026}

\title{SinGeo: Unlock Single Model's Potential for Robust Cross-View Geo-Localization}

\author{
	Yang Chen$^{1}$,
	Xieyuanli Chen$^{1}$\thanks{Corresponding authors. \newline \quad Emails: wutao@nudt.edu.cn, xieyuanli.chen@nudt.edu.cn},
	Junxiang Li$^{1}$,
	Jie Tang$^{1}$,
	Tao Wu$^{1}$\footnotemark[1]
	\\
	$^{1}$College of Intelligence Science and Technology, National University of Defense Technology
}
\begin{document}

\maketitle

\begin{abstract}
	
	
	Robust cross-view geo-localization (CVGL) remains challenging despite the surge in recent progress. Existing methods still rely on field-of-view (FoV)-specific training paradigms, where models are optimized under a fixed FoV but collapse when tested on unseen FoVs and unknown orientations. This limitation necessitates deploying multiple models to cover diverse variations. Although studies have explored dynamic FoV training by simply randomizing FoVs, they failed to achieve robustness across diverse conditions---implicitly assuming all FoVs are equally difficult. To address this gap, we present SinGeo, a simple yet powerful framework that enables a \textbf{Sin}gle model to realize robust cross-view \textbf{Geo}-localization without additional modules or explicit transformations. SinGeo employs a dual discriminative learning architecture that enhances intra-view discriminability within both ground and satellite branches, and is the first to introduce a curriculum learning strategy to achieve robust CVGL. Extensive evaluations on four benchmark datasets reveal that SinGeo sets state-of-the-art (SOTA) results under diverse conditions, and notably outperforms methods specifically trained for extreme FoVs. Beyond superior performance, SinGeo also exhibits cross-architecture transferability. Furthermore, we propose a consistency evaluation method to objectively assess model stability under varying views, providing an objective perspective for understanding and advancing robustness in future CVGL research. Codes are available at: \url{https://github.com/Yangchen-nudt/SinGeo}.

\end{abstract}

\section{Introduction}
\label{sec:intro}
\begin{figure}[t]
	\centering
	\includegraphics[width=\linewidth]{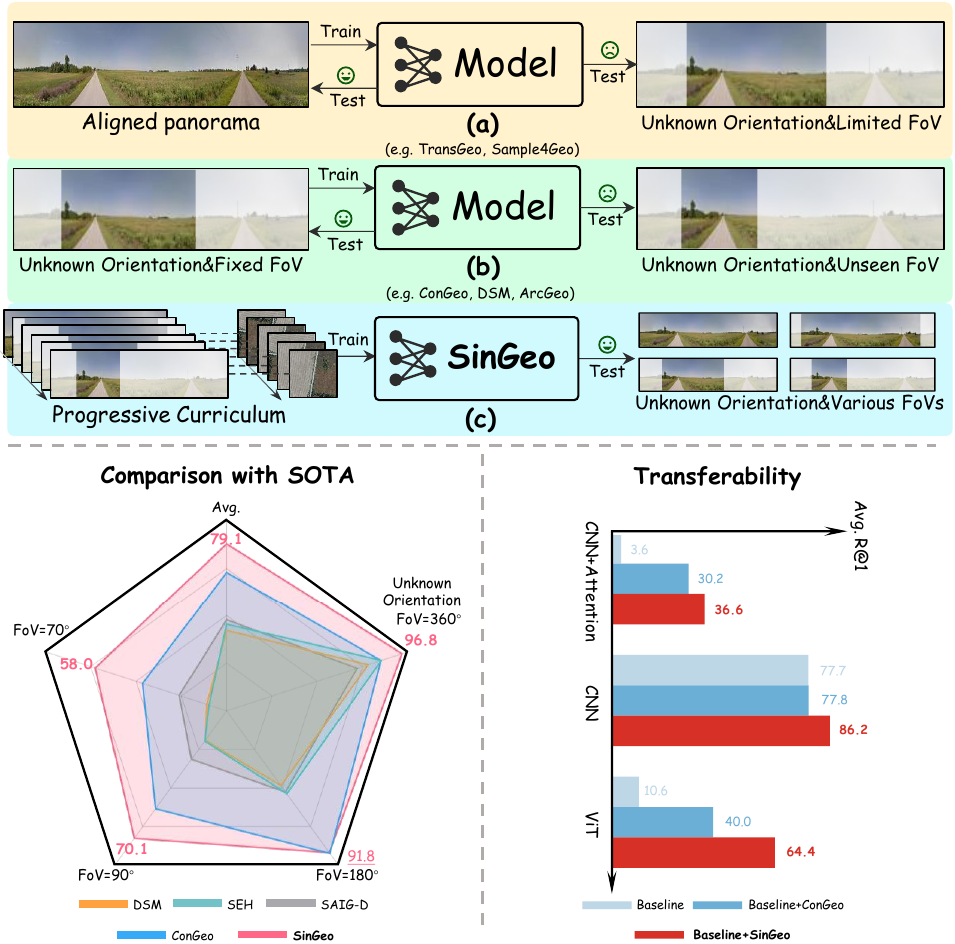}
	\caption{{Comparison of different training paradigms for Cross-view Geo-localization.} {SinGeo achieves superior overall performance and better transferability under scenarios of both unknown orientation and different limited field of views. Previous methods including DSM \cite{dsm}, SEH \cite{seh}, SAIG-D \cite{saig}, and ConGeo \cite{congeo} are reported with their Top-1 Recall performance on CVUSA \cite{cvusa} dataset.}}
	\label{fig:overview}
\end{figure}
Cross-view geo-localization (CVGL) is a geographic location technology by cross-view matching ground-view query image with Geo-tagged satellite database images \cite{lin2013cross,cvm}. This task holds broad application potential in robotic navigation \cite{li2019navi}, autonomous driving and augmented reality \cite{dehi}. 

Traditional CVGL approaches \cite{transgeo,geodtr,sample4geo} have mainly focused on idealized benchmarks \cite{cvusa,cvact} where ground-view images are north-aligned panoramas, and performance on these datasets has largely saturated. However, this paradigm limits the practical utility of CVGL \cite{sample4geo} as shown in Fig.~\ref{fig:overview}(a). In real-world practice, ground images are typically captured by consumer-grade devices such as smartphones or vehicle cameras \cite{congeo}. These images have unknown orientations and restricted FoV, often ranging from 70\textdegree to 180\textdegree. Such discrepancies introduce significant domain gaps, making robustness and generalization crucial for the wide adoption of CVGL in practical scenarios.

Although recent studies \cite{dsm, gal, congeo} have made progress in robust CVGL, existing methods still face key limitations. Some rely on explicit view transformations, such as polar transformation \cite{dsm, arcgeo} or bird’s-eye-view projections to reduce the cross-view discrepancy \cite{bevcv}. While simplifying the task via unified perspective learning, these designs inevitably introduce image distortions and depend on predefined parameters \cite{bevcv}. More recent robust CVGL methods \cite{arcgeo, dehi, gal} tackle the problem from the perspective of data augmentation, generating fixed-FoV samples from panoramas \cite{congeo} and refining architectures or training strategies. However, such a FoV-specific training paradigm leads to good performance only at the trained FoV, and degrades drastically on unseen FoVs as shown in Fig.~\ref{fig:overview}(b). In this context, different models are required for different configurations. These observations raise an important research question: without explicit transformations or additional modules, \textit{can a single model inherently achieve consistent and high performance under varying orientations and  FoVs?}

To answer it, we propose SinGeo, a simple yet effective and transferable framework that enhances a single model’s robustness across diverse viewing conditions. SinGeo integrates two synergistic ideas: 1) a dual discriminative learning architecture that strengthens intra-view discriminability for each branch; 2) a progressive training strategy inspired by curriculum learning \cite{cl} as shown in Fig.~\ref{fig:overview}(c), that schedules learning difficulty to guide the model from simple to more challenging scenarios.  SinGeo consistently achieves strong and balanced performance across varying  FoVs and orientations, surpassing previous  FoV-specific approaches even under extreme  FoV conditions.

To further investigate the mechanism behind improved robustness, we quantitatively measure how consistent a model remains under orientation and FoV variations. This evaluation provides a perspective for understanding and advancing robustness in future CVGL research. We summarize our contributions as follows:

\begin{itemize}
	\item We present SinGeo, a pratical solution that delivers a paradigm shift to learning a single model for robust CVGL via effective combination of proposed dual discriminative learning and curriculum-guided progressive training. It sets SOTA results in multiple benchmarks, significantly outperforming FoV-specific methods under extreme FoVs.
	\item SinGeo can be integrated into multiple  CVGL backbones, including ViT \cite{vit}, CNN \cite{convnext}, and CNN+attention \cite{geodtr}, consistently enhancing their robustness and outperforming another plug-and-play method, ConGeo \cite{congeo}.
	\item We provide a new evaluation perspective. Through quantitative consistency analysis, we provide valuable insights into the robustness and stability of CVGL models, offering a potential direction for future research to measure robustness in CVGL.
\end{itemize}


\section{Related work}
\label{sec:related_work}
\begin{figure*}[t]
	\centering
	\includegraphics[width=\linewidth]{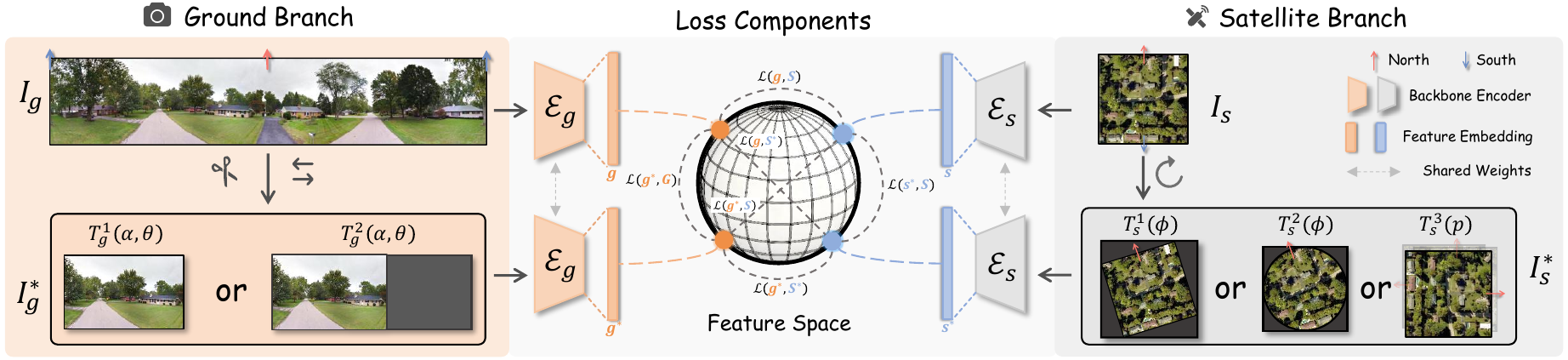}
	\caption{The proposed dual discriminative learning architecture. “Dual” denotes the two-branch design that enhances intra-view discriminativeness through self-supervision in both ground and satellite branches. Specifically, each branch applies specific transformations to generate \(I_g^*\) and \(I_s^*\).  The learning objective combines contrastive losses for intra-view discrimination and cross-view alignment.}
	\label{fig:dualloss}
\end{figure*}

\paragraph{Traditional Cross-View Geo-Localization.} 
In CVGL studies, a common paradigm to perform cross-view matching \cite{cvm,cvft} is extracting feature embeddings independently from ground-view and satellite images, and then aligning their similarities through various contrastive learning loss functions \cite{arcgeo, cvusa, softmargin, sample4geo, seh}. Early approaches attempted to explicitly bridge the cross-view gap. Shi \etal \cite{safa,dsm} converted satellite images into pseudo ground-view representations via polar transformation. Alternatively, some methods \cite{epbev,bevcv} projected ground-view imagery or intermediate features into bird’s-eye view.

More recent research has shifted toward transformation-free architectures that align representations through advanced network design. TransGeo \cite{transgeo} proposed the first pure transformer-based method achieving better performance. Besides, hybrid CNN–Attention frameworks \cite{geodtr,safa} leveraging local–global feature extraction further improved representation ability. Notably, Sample4Geo \cite{sample4geo} used a single CNN encoder with InfoNCE loss \cite{infonce} and proposed new hard negative sampling strategies, reaching competitive performance without additional modules or transformations.

As also discussed in \cite{sample4geo}, benchmarks such as CVUSA \cite{cvusa} and CVACT \cite{cvact} with north-aligned and center-aligned panoramas are becoming saturated. Previous methods collapse severely under practical conditions where images have unknown orientation and narrow FoV \cite{congeo}.

\paragraph{Robust Cross-View Geo-Localization.}
 Robust CVGL focuses on maintaining performance under unknown orientation and limited FoV conditions. Shi \etal \cite{dsm} pioneered this line of work with the DSM module, which uses a fixed sliding window over the query-activated region for orientation estimation and robust matching. Follow-up studies mainly explored strategies based on data augmentation. For scenarios with unknown orientation, DeHi \cite{dehi} contrasted representations of polar-transformed satellite images and unaligned panoramas. For scenarios with both unknown orientation and limited FoVs, models such as GAL \cite{gal} and ArcGeo \cite{arcgeo} randomly shifted and cropped panoramas into fixed FoV patches, substituting them for the original panoramic inputs to strengthen their limited-FoV representations. Specifically, ConGeo \cite{congeo} provided a novel approach to align embeddings from panoramas and their cropped ones under particular FoVs, thus enhancing its performance in the corresponding limited-FoV occasions. 

However, existing studies focus more on the ground branch, overlooking the satellite one. More noteworthy, most of them train their models solely at one particular FoV. While effective at the trained FoV, these models collapse when tested at unseen ones. Developing a method or even a paradigm that performs uniformly across limited FoVs remains a crucial and unsettled endeavor.

\paragraph{Curriculum Learning.}
Curriculum Learning (CL) \cite{cl}, first proposed by Bengio \etal with convergence proof, was inspired by the human learning progress from easy to difficult tasks. Li \etal \cite{clcvgl} apply CL to the unsupervised learning paradigm in traditional CVGL. Yet, CL remains underexplored in robust CVGL. Few studies explore the potential of training with dynamic FoVs. ConGeo \cite{congeo} and TransGeo \cite{transgeo} tried training models with 0\textdegree--360\textdegree random FoVs, but underperformed their fixed-FoV setups. This design implicitly assumes all FoVs are equally difficult. Our work aims to question this assumption by introducing a CL-inspired strategy that enables a single model to perform effectively and consistently at different FoVs.

\section{SinGeo}
\label{sec:SinGeo}

To unleash a single model’s potential for robust CVGL, we propose SinGeo, a dual discriminative learning architecture coupled with a curriculum learning strategy. Notably, both of them are module-free, granting the integration of SinGeo into different CVGL backbones.


\subsection{Dual Discriminative Learning}
\label{sec:3.1}

Traditional CVGL methods mainly align cross-view features via contrastive losses \cite{infonce, softmargin, triplet}, but objectives like InfoNCE \cite{infonce} can lead to shortcut solutions \cite{shortcut}. Although prior work \cite{congeo, gal} improves single-branch representations, a unified design equally enhancing both branches remains underexplored. Under limited FoV and unknown orientation, strong intra-view discriminability is crucial for extracting meaningful features.

\paragraph{Contrastive Learning for Discriminativeness.} We denote the ground-view image as \(I_g\) and the satellite image as \(I_s\), with their corresponding encoders \(\mathcal{E}_g\) and \(\mathcal{E}_s\).

To enhance the discriminativeness of the model, we first generate positive samples \(I_g^*\) with unknown orientation and limited FoV, following previous studies \cite{congeo, arcgeo}. This approach aims to create new positive sample pairs \((I_g, I_g^*)\), reducing the feature distance between \(I_g\) and \(I_g^*\) so that they can be aligned with \(I_s\) in a more unified feature space. 

In fact, we observe that another feasible way to create sample pairs under unaligned orientation is to rotate the satellite image \(I_s\) to \(I_s^*\), thereby generating another set of positive sample pairs \((I_s, I_s^*)\). This design essentially enables the model to focus on discriminative regions in \(I_s\) through self-supervision, rather than merely learning the correspondence between \(I_s\) and \(I_g\).

This dual design as shown in Figure~\ref{fig:dualloss} allows the model to focus on both the satellite branch and the ground branch themselves, and prevents the model from “bias” such as over-focusing on a single branch, or blindly searching for corresponding parts between satellite images and ground view images during training while forgetting to extract features of truly important regions in each branch. 

Specifically, the loss for discriminativeness is given by

\begin{subequations}
	\begin{equation}
		\mathcal{L}(g^*, G) = 
		-\log \frac{\exp\left(g^* \cdot g_+ / \tau\right)}
		{\sum_{g_i \in G} \exp\left(g^* \cdot g_i / \tau\right)},
		\tag{1a}\label{eq:info_g}
	\end{equation}
	\begin{equation}
		\mathcal{L}(s^*, S) = 
		-\log \frac{\exp\left(s^* \cdot s_+ / \tau\right)}
		{\sum_{s_i \in S} \exp\left(s^* \cdot s_i / \tau\right)},
		\tag{1b}\label{eq:info_s}
	\end{equation}
	\begin{equation}
		\mathcal{L}_{\text{disc}} = 
		\mathcal{L}(g^*, G) + \mathcal{L}(s^*, S).
		\tag{2}\label{eq:l_robust}
	\end{equation}
\end{subequations}


\( g \) and \(g^*\) denote the encoded embeddings of \(I_g\) and  \(I_g^*\), respectively. \( G \) represents the set of embeddings encoded from a collection of \(\{ I_g \}\), where \(g_+ \in G\) is the only positive sample matching \(g^*\), and the rest of \( G \) are all negative samples.  \( s \), \(s^*\), \(s_+\), and \( S \) are defined analogously.

As illustrated in Fig.~\ref{fig:dualloss}, we utilize several transformation alternatives to generate \( I_g^* \) and \( I_s^* \). For the ground-view branch, two distinct transformations \( T_g^1(\alpha, \theta) \) and \( T_g^2(\alpha, \theta) \) are applied. For CNN-based models, \( T_g^1 \) randomly shifts the panorama horizontally by an angle \(\alpha\) and crops a view with FoV \(\theta\). For ViT-based models, \( T_g^2 \) further pads the cropped view with zeros into the original panorama size. For the satellite branch, we consider two continuous rotation transformations, \( T_s^1(\phi) \) and \( T_s^2(\phi) \), and a discrete one \( T_s^3(p) \). Here, \(\phi\) denotes the angle of yaw rotation. \( T_s^1(\phi) \) and \( T_s^2(\phi) \) perform outer-- and inner--bounding rotations, respectively. \( T_s^3(p) \) applies a clockwise rotation by one of \( \{90^\circ, 180^\circ, 270^\circ\} \) with probability \( p \). All transformations additionally include color-level augmentations, such as variations in brightness and saturation.

\paragraph{Contrastive Learning for Cross-view Alignment.} 
After collecting a richer set of cross-view samples \((I_g, I_s, I_g^*, I_s^*)\), we introduce the cross-view learning loss of SinGeo to perform feature alignment between different views, defined as

\begin{equation}
	\begin{aligned}
		\mathcal{L}_{\text{cross}} =\;
		& \mathcal{L}(g, S)
		+ \omega_1 \mathcal{L}(g^*, S) \\
		& + \omega_2 \mathcal{L}(g, S^*)
		+ \omega_3 \mathcal{L}(g^*, S^*),
	\end{aligned}
	\tag{3}\label{eq:crossloss}
\end{equation}

where each loss term \(\mathcal{L}(\cdot, \cdot)\) follows the similar definition in Eq.~(\ref{eq:info_g})--(\ref{eq:info_s}), and \(\omega_1, \omega_2, \omega_3\) are balancing weights of the augmented sample pairs. Finally, the training objective of SinGeo is

\begin{equation}
	\mathcal{L}_{\text{total}} = 
	\mathcal{L}_{\text{cross}} 
	+ \gamma \mathcal{L}_{\text{disc}},
	\tag{4}\label{eq:totalloss}
\end{equation}

where \(\gamma\) is a trade-off factor that balances discriminativeness and cross-view alignment.

\begin{figure*}[htbp]
	\centering
	\includegraphics[width=\linewidth]{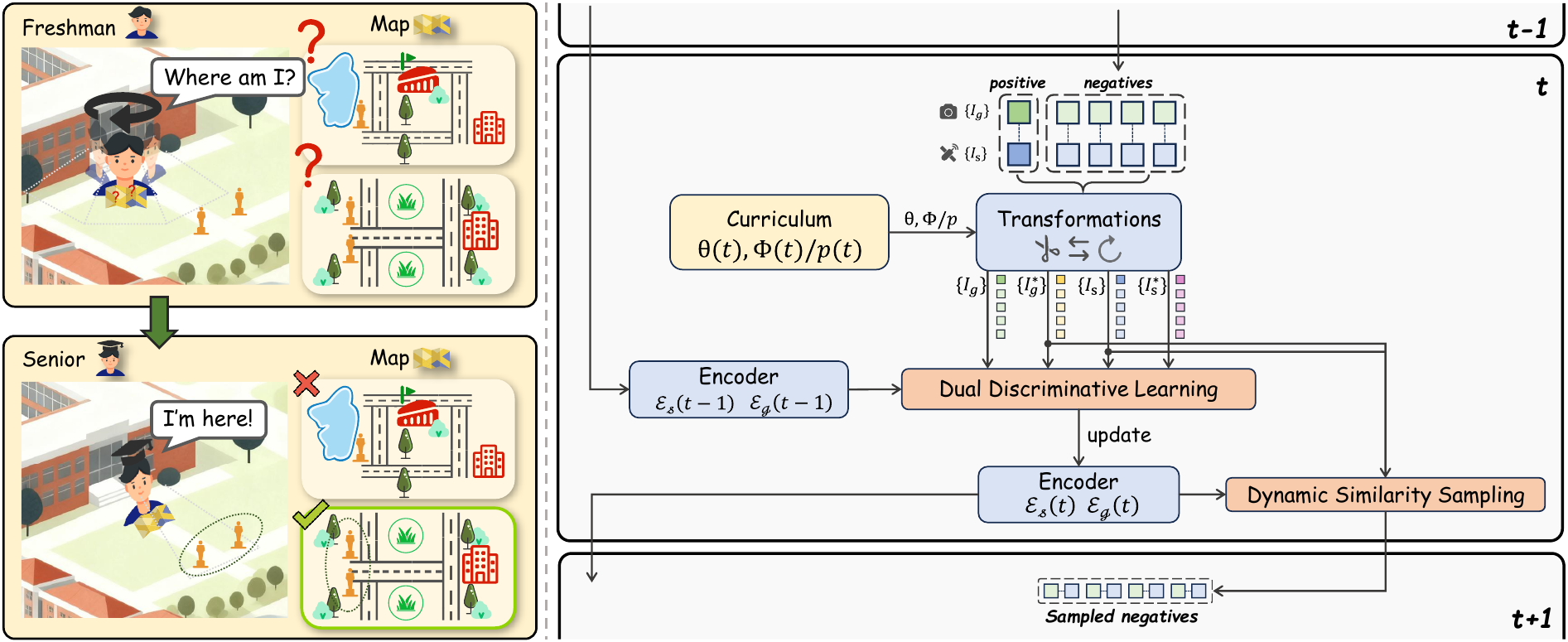}
	\caption{{Illustration of the curriculum learning framework and its inspiration.} {Left}: An intuitive analogy of curriculum learning, where a freshman gradually refines geo-localization ability when he gets more familiar and practiced. {Right}:. The predetermined curriculum schedules the difficulties according to the epoch $t$. After updated by dual discriminative learning, the encoder is then fed into Dynamic Similarity Sampling block \cite{sample4geo} to generate negatives for next epoch.}
	\label{fig:curriculum}
\end{figure*}

\begin{figure*}[t]
	\centering
	\includegraphics[width=\linewidth]{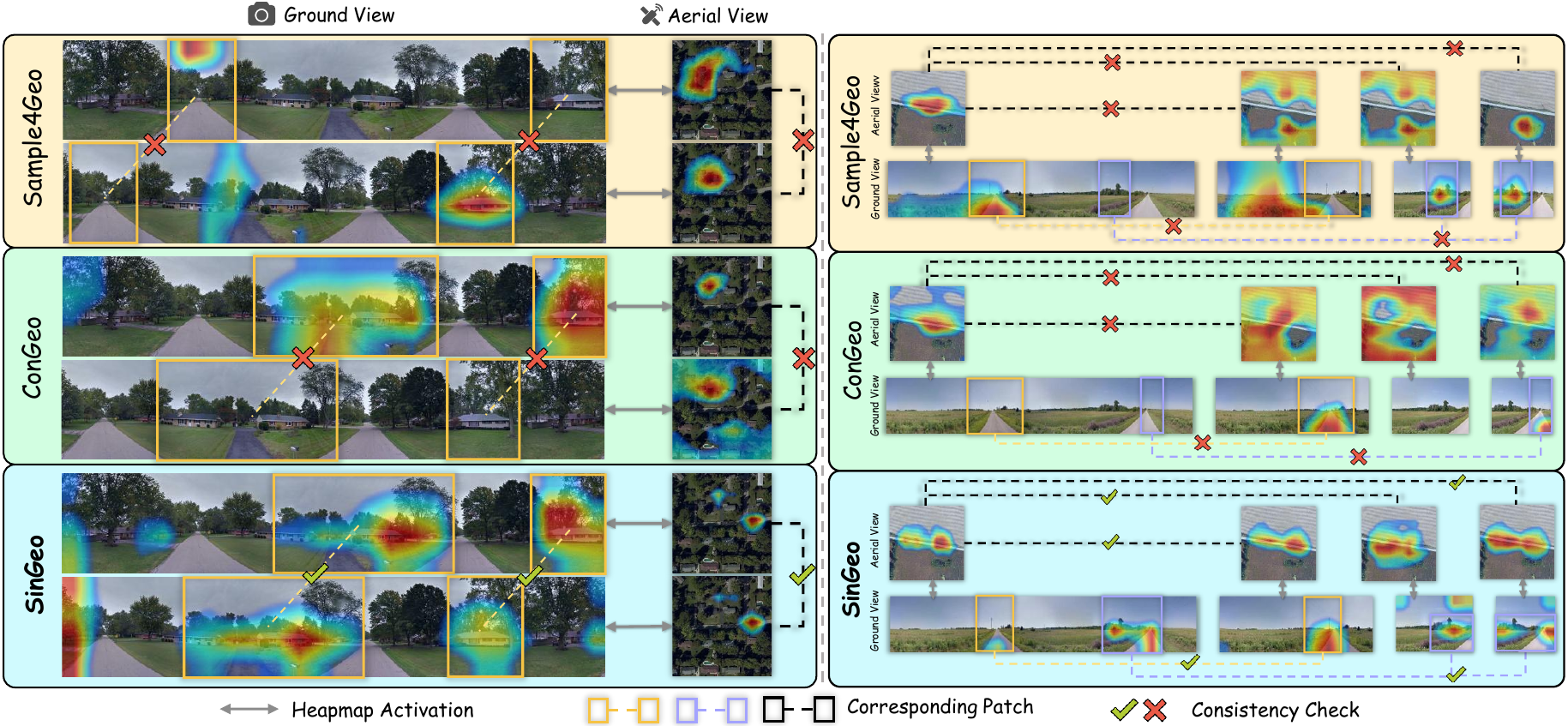}
	\caption{Visualization of consistency comparison among SinGeo, ConGeo and Sample4Geo on the CVUSA dataset. The left column depicts a sample under varying orientations, where SinGeo exhibits stronger consistency in both ground and satellite branches. The right column presents a sample under different FoVs, where the activation heatmaps of SinGeo remain stable across varying views.}
	\label{fig:consistency}
\end{figure*}

\subsection{Curriculum Learning for Robust CVGL}
\label{sec:3.2}

In Fig.~\ref{fig:curriculum}, we recall how human learn cross-view geo-localization in everyday experience. When inexperienced or unfamiliar with a location, individuals initially conduct a 360\textdegree panoramic scan and then match the visual context with a map to infer their position. With increased familiarity or experience, they can pinpoint their location on the map via a limited FoV instead of panoramic observation. 

Inspired by this observation and curriculum learning, we enable the CVGL model to learn progressively. As illustrated in Fig.~\ref{fig:curriculum}, the proposed curriculum learning (CL) process dynamically modulates the generation of \((I_g, I_s, I_g^*, I_s^*)\) during training according to the current epoch \(t\). After each update via backpropagation of the dual discriminative learning loss, the updated encoders \(\mathcal{E}_g(t)\) and \(\mathcal{E}_s(t)\) perform Dynamic Similarity Sampling~\cite{sample4geo} on \(I_g^*\) and \(I_s\) to mine new sampled negatives for the next epoch based on visual similarity. 

To realize this progressive training process, we dynamically adjust $T_g(\alpha, \theta)$ and $T_s(\phi)$ or $T_s(p)$ to enable an easy-to-hard transition over epochs. Specifically, let the total training epochs be $ n $, and the current epoch be $ t $. To achieve this, we parameterize the dynamic evolution of the parameters $\theta$, $\phi$, and $p$ as functions of the training progress $ t/n $. 

Formally, let $\eta \in \{\theta, \phi, p\}$ denote a generic parameter, with initial value $\eta_{\text{init}}$ and final value $\eta_{\text{final}}$ corresponding to an easy beginning and a hard ending respectively. 

\begin{equation}
	\eta(t) = \eta_{\text{init}} + (\eta_{\text{final}} - \eta_{\text{init}}) \cdot f(t/n), \tag{5}\label{dynamicfunc}
\end{equation}

where \(f(\cdot)\) is a monotonically increasing scheduling function that adjusts the transition pace based on training progress \(t/n\). For $\theta$ where $\eta_{\text{init}} > \eta_{\text{final}}$, this formulation yields a reduction in $\theta$ from the easy to hard stage. Reversely, for $\phi$ and $p$ where $\eta_{\text{init}} < \eta_{\text{final}}$, it results in an increase in $\phi/p$ to transition from easy to hard. We consider three variants of $ f $:

\begin{itemize}
	\item Linear: \(f_1(x) = x\), corresponding to a uniform progression from easy to hard.
	\item Exponential (fast-to-slow): \(f_2(x) = \frac{1 - \exp(-\lambda x)}{1 - \exp(-\lambda)}\), resulting in rapid initial changes that slow down gradually. $\lambda > 0$ controls the decay rate.
	\item Exponential (slow-to-fast): \(f_3(x) = \frac{\exp(\lambda x) - 1}{\exp(\lambda) - 1}\), yielding slow initial changes but accelerating over time.
\end{itemize}

The three variants reflect different patterns of human learning. Some learners improve steadily, others learn quickly then refine gradually, or start slowly and accelerate later. A comparative analysis of these scheduling functions will be presented in our experiments.


\section{Consistency: An Explainable and Quantitative Perspective}
\label{sec:analysis}
To understand the reasons behind SinGeo’s superior performance over existing methods in Fig.~\ref{fig:overview}, we provide a visual explanation from the perspective of consistency through the lens of Grad-CAM \cite{gradcam}. The activation maps in Fig.~\ref{fig:consistency} show that SinGeo maintains stable activated regions in both satellite and ground branches when the orientation or FoV changes. In contrast, previous methods exhibit dispersed or shifting activation. This observation suggests that SinGeo achieves more consistent cross-view matching, enabling it to generalize effectively beyond north-aligned panorama scenarios.

To explicitly formalize the notion of consistency, we obtain the activation heatmap \(\mathcal{H}\) of a query embedding \({e}_{\text{query}}\) on its corresponding reference image \(I_{\text{refer}}\) using Grad-CAM \cite{gradcam}, defined as:
\[
\mathcal{H} = \text{GradCAM}({e}_{\text{query}}, I_{\text{refer}}).
\]

For a random orientation \(\alpha\) and a given FoV \(\theta\), the transformed ground-view image is denoted as \(I_g^*(\theta) = T_g(I_g \mid \alpha, \theta)\). Correspondingly, \({g}\) and \({g}^*(\theta)\) are the embeddings of \(I_g\) and \(I_g^*(\theta)\) encoded by \(\mathcal{E}_g\), while \({s}\) is the embedding of \(I_s\) encoded by the satellite encoder \(\mathcal{E}_s\). We derive four heatmaps:

\[
\left\{
\begin{aligned}
	&\mathcal{H}_s = \text{GradCAM}(g, I_s) \\
	&\mathcal{H}_s^*(\theta) = \text{GradCAM}(g^*(\theta), I_s) \\
	&\mathcal{H}_g = \text{GradCAM}(s, I_g) \\
	&\mathcal{H}_g^*(\theta) = \text{GradCAM}(s, I_g^*(\theta))
\end{aligned}
\right.
\tag{6}
\]

We apply the same transformation \({T}_g(\alpha, \theta)\) to \(\mathcal{H}_g\), written as \({T}_g(\mathcal{H}_g\mid\alpha, \theta)\). A robust model should ensure that \(\mathcal{H}_s\) and \(\mathcal{H}_s^*(\theta)\) remain consistent in the satellite branch, and that \({T}_g(\mathcal{H}_g\mid\alpha, \theta)\) and \(\mathcal{H}_g^*(\theta)\) are also consistent in the ground branch. To quantify this consistency, we adopt the normalized Structural Similarity Index (SSIM) \cite{ssim} and define two consistency metrics, orientation-consistency (OC) and FoV-consistency (FC), as follows:

\begin{equation}\tag{7}
	\footnotesize
	\left\{
	\begin{aligned}
		\text{OC}_\text{sat} &= \text{SSIM}(\mathcal{H}_s, \mathcal{H}_s^*(360^\circ)) \\
		\text{OC}_\text{grd} &= \text{SSIM}\!\left( {T}_g(\mathcal{H}_g \mid \alpha, 360^\circ), \mathcal{H}_g^*(360^\circ) \right) \\
		\text{FC}_\text{sat} &= \frac{1}{3} \sum_{\theta_i \in \{180^\circ, 90^\circ, 70^\circ\}} \text{SSIM}(\mathcal{H}_s, \mathcal{H}_s^*(\theta_i)) \\
		\text{FC}_\text{grd} &= \frac{1}{3} \sum_{\theta_i \in \{180^\circ, 90^\circ, 70^\circ\}} \text{SSIM}\!\left( {T}_g(\mathcal{H}_g \mid \alpha, \theta_i), \mathcal{H}_g^*(\theta_i) \right)
	\end{aligned}
	\right.
\end{equation}

\begin{table*}[t]
	
	\centering
	\small
	\setlength{\tabcolsep}{1.5pt}  
	\definecolor{graybg}{rgb}{0.9, 0.9, 0.9}  
	\caption{Comparison with single-model performance of state-of-the-art methods on CVUSA and CVACT Val datasets under unknown orientation and limited FoV settings. The results of methods with multiple FoV-specialized trained models are reported as reference with \colorbox{graybg}{gray background}. \textbf{Bold} indicates the best performance among single-model methods . \underline{Underline} denotes the best performance across all methods. “–” denotes metrics not provided in the original paper.}
	\begin{tabular}{@{}l|l|c|cccc|cccc|cccc|cccc@{}}
		\toprule  
		\multirow{2}{*}{\centering Set} & \multirow{2}{*}{\centering Methods} 
		& \multirow{1}{*}{\centering \makecell{Avg.}}  
		& \multicolumn{4}{c|}{\makecell{FoV $360^\circ$}} 
		& \multicolumn{4}{c|}{\makecell{FoV $180^\circ$}} 
		& \multicolumn{4}{c|}{\makecell{FoV $90^\circ$}} 
		& \multicolumn{4}{c}{\makecell{FoV $70^\circ$}} \\
		& & R@1 & R@1 & R@5 & R@10 & R@1\% & R@1 & R@5 & R@10 & R@1\% & R@1 & R@5 & R@10 & R@1\% & R@1 & R@5 & R@10 & R@1\% \\
		\midrule  
		\multirow{11}{*}{\rotatebox{90}{CVUSA}}  
		& CVM-Net \cite{cvm}        & 7.3  & 16.3 & 38.9 & 49.4 & 88.1 & 7.4  & 22.5 & 32.6 & 75.4 & 2.8  & 10.1 & 16.7 & 55.5 & 2.6  & 9.3  & 15.1 & 21.8 \\
		& CVFT \cite{cvft}           & 10.0 & 23.4 & 44.4 & 55.2 & 86.6 & 8.1  & 24.3 & 34.5 & 75.2 & 4.8  & 14.8 & 23.2 & 61.2 & 3.8  & 12.4 & 19.3 & 55.6 \\
		& SEH \cite{seh}            & 40.9    & 85.4 & 93.5 & 95.8 & -    & 53.7 & 72.3 & 79.0 & -    & 16.6 & 32.2 & 40.3 & -    & 7.8  & 18.8 & 25.6 & -    \\
		& SAIG-D \cite{saig}         & 43.0 & 72.0 & 90.2 & 94.0 & 99.1 & 52.5 & 78.1 & 85.8 & 97.7 & 26.7 & 50.2 & 59.8 & 86.6 & 20.9 & 41.4 & 51.2 & 80.4 \\
		& GAL \cite{gal}            & -    & -    & -    & -    & -    & 48.9 & 69.9 & 78.5 & 95.7 & 22.5 & 44.4 & 54.2 & 83.6 & 15.2 & 32.9 & 42.1 & 75.2 \\
		& DeHi \cite{dehi}          & -    & 82.4 & 93.5 & 96.3 & 99.4 & 60.4 & 81.8 & 88.2 & 98.0 & 31.5 & 55.1 & 65.6 & 90.8 & -    & -    & -    & -    \\
		& ArcGeo \cite{arcgeo}        & -    & -    & -    & -    & -    & 63.7 & 85.4 & 90.6 & -    & 44.2 & 70.3 & 78.8 & -    & 37.7 & 64.4 & 73.6 & -    \\
		& Sample4Geo \cite{sample4geo}     & 68.5 & 93.3 & 97.5 & 98.0 & 99.1 & 84.6 & 95.9 & 97.6 & \textbf{99.5} & 55.1 & 78.3 & 85.0 & 96.6 & 40.9 & 65.4 & 74.1 & 93.0 \\
		& \textbf{SinGeo} & \underline{\textbf{79.1}} & \underline{\textbf{96.8}} & \underline{\textbf{99.0}} & \underline{\textbf{99.2}} & {\textbf{99.5}} & {\textbf{91.8}} & {\textbf{97.6}} & {\textbf{98.4}} & {\textbf{99.5}} & \underline{\textbf{70.1}} & \underline{\textbf{86.6}} & \underline{\textbf{91.0}} & \underline{\textbf{97.5}} & \underline{\textbf{58.0}} & \underline{\textbf{78.1}} & \underline{\textbf{83.9}} & \underline{\textbf{95.2}} \\
		& \cellcolor{graybg}DSM \cite{dsm}      & \cellcolor{graybg}37.9 & \cellcolor{graybg}78.1 & \cellcolor{graybg}89.5 & \cellcolor{graybg}92.9 & \cellcolor{graybg}98.5 & \cellcolor{graybg}48.5 & \cellcolor{graybg}68.5 & \cellcolor{graybg}75.6 & \cellcolor{graybg}93.0 & \cellcolor{graybg}16.2 & \cellcolor{graybg}31.4 & \cellcolor{graybg}39.9 & \cellcolor{graybg}71.1 & \cellcolor{graybg}8.8  & \cellcolor{graybg}19.9 & \cellcolor{graybg}27.3 & \cellcolor{graybg}61.2 \\
		& \cellcolor{graybg}ConGeo \cite{congeo} & \cellcolor{graybg}73.4 & \cellcolor{graybg}96.6 & \cellcolor{graybg}98.9 & \underline{\cellcolor{graybg}99.2} & \underline{\cellcolor{graybg}99.7} & \underline{\cellcolor{graybg}92.3} & \underline{\cellcolor{graybg}97.9} & \underline{\cellcolor{graybg}98.7} & \underline{\cellcolor{graybg}99.7} & \cellcolor{graybg}55.5 & \cellcolor{graybg}75.4 & \cellcolor{graybg}81.5 & \cellcolor{graybg}93.9 & \cellcolor{graybg}49.1 & \cellcolor{graybg}70.8 & \cellcolor{graybg}78.0 & \cellcolor{graybg}93.1 \\
		\midrule  
		\multirow{8}{*}{\rotatebox{90}{CVACT}}  
		& CVM-Net \cite{cvm}       & 4.9  & 13.1 & 33.9 & 45.7 & 81.8 & 3.9  & 13.7 & 21.2 & 59.2 & 1.5  & 5.7  & 9.6  & 38.1 & 1.2  & 5.0  & 8.4  & 34.7 \\
		& CVFT \cite{cvft}          & 10.9 & 26.8 & 46.9 & 55.1 & 81.0 & 8.1  & 24.3 & 34.5 & 75.2 & 4.8  & 14.8 & 23.2 & 61.2 & 3.8  & 12.4 & 19.3 & 55.6 \\
		& SEH \cite{seh} &36.5 &77.4 &88.6 &90.9 &- &47.7& 67.9 & 74.3 &- &13.9 &28.4 &36.2 &- &6.9 &16.5 &22.3 &- \\
		& GAL \cite{gal} &- &- &- &- &- &49.9 &68.5 &77.2 &93.0 &26.1 &49.2 &59.3 &85.6 &14.2 &33.0 &43.2 &77.2 \\
		& Sample4Geo \cite{sample4geo}    & 47.0 & 82.4 & 90.6 & 92.3 & 95.4 & 58.9 & 79.8 & 85.3 & \textbf{95.0} & 27.9 & 52.0 & 62.3 & 87.0 & 18.8 & 40.4 & 51.0 & 81.3 \\
		& \textbf{SinGeo} & \underline{\textbf{55.8}} & \underline{\textbf{83.4}} & \underline{\textbf{91.1}} & \underline{\textbf{92.6}} & {\textbf{95.7}} & {\textbf{68.0}} & {\textbf{82.4}} & {\textbf{85.8}} & 93.0 & \underline{\textbf{42.6}} & \underline{\textbf{65.5}} & \underline{\textbf{72.8}} & \underline{\textbf{89.0}} & \underline{\textbf{29.0}} & \underline{\textbf{51.8}} & \underline{\textbf{61.0}} & \underline{\textbf{83.4}} \\
		& \cellcolor{graybg}DSM \cite{dsm}      & \cellcolor{graybg}37.1 & \cellcolor{graybg}72.9 & \cellcolor{graybg}85.7 & \cellcolor{graybg}88.9 & \cellcolor{graybg}95.3 & \cellcolor{graybg}49.1 & \cellcolor{graybg}67.8 & \cellcolor{graybg}74.2 & \cellcolor{graybg}89.9 & \cellcolor{graybg}18.1 & \cellcolor{graybg}33.3 & \cellcolor{graybg}40.9 & \cellcolor{graybg}68.7 & \cellcolor{graybg}8.3  & \cellcolor{graybg}20.7 & \cellcolor{graybg}27.1 & \cellcolor{graybg}57.1 \\
		& \cellcolor{graybg}ConGeo \cite{congeo} & \cellcolor{graybg}54.6 & \cellcolor{graybg}83.0 & \cellcolor{graybg}90.6 & \cellcolor{graybg}92.4 & \underline{\cellcolor{graybg}96.3} & \underline{\cellcolor{graybg}70.3} & \underline{\cellcolor{graybg}85.2} & \underline{\cellcolor{graybg}88.6} & \underline{\cellcolor{graybg}95.1} & \cellcolor{graybg}40.6 & \cellcolor{graybg}62.6 & \cellcolor{graybg}69.8 & \cellcolor{graybg}86.6 & \cellcolor{graybg}24.6 & \cellcolor{graybg}45.3 & \cellcolor{graybg}54.3 & \cellcolor{graybg}80.6 \\
		\bottomrule  
	\end{tabular}

	\label{tab:fov_comparison}
\end{table*}

\section{Experiments}

\label{sec:Experimental}
\subsection{Experimental Setup}
\paragraph{Datasets.}
We conduct experiments on four datasets. CVUSA \cite{cvusa} and CVACT \cite{cvact} are two standard datasets for CVGL, featuring north-aligned, center-aligned panoramas and satellite images. CVUSA includes 35,532 image pairs for training and 8,884 for evaluation. CVACT has the same size for its training and validation sets, with an additional test set containing 92,802 pairs. VIGOR \cite{vigor} is a challenging non-center-aligned dataset, containing 90,618 aerial-view and 105,214 ground-view images across four cities (New York, Seattle, San Francisco, Chicago). It uses two splits, namely same-area and cross-area: same-area uses all cities for training and validation, while cross-area conducts training on New York and Seattle and testing on San Francisco and Chicago for generalization evaluation. University-1652 \cite{university} is a difficult CVGL dataset, containing ground-view images with unknown orientations and a limited FoV to evaluate models in real-world scenarios. Additionally, University-1652 offers a limited amount of data for training. For street-to-satellite matching, it only includes 2,579 street images and 951 satellite images, and for satellite-to-street matching, it comprises 701 satellite images and 2,921 street images.

\paragraph{Metrics.} 
Following prior methods, we report retrieval performance using Top-$k$ recall, denoted as R@$k$. For each query ground image, we retrieve the $k$ nearest reference neighbors in the embedding space based on cosine similarity. A retrieval is correct if the ground-truth reference appears in the top $k$. Besides, we also report the new consistency metrics proposed in Sec.~\ref{sec:analysis}. In the main paper, we report SSIM-based consistency, while \textit{supplementary material} includes consistency results based on cosine similarity and pearson correlation coefficient \cite{pcc}.

\paragraph{Implementation Details.} 
In the main experiments, we use ConvNeXt-B as our backbone. The weights $\omega_1$, $\omega_2$, $\omega_3$, and $\gamma$ are set to 0.25, 0.25, 0.25, and 0.5 empirically. Label smoothing of 0.1 is used in the InfoNCE loss. To generate ground-view images with unknown orientation and limited FoV, \(T_{g}^{1}(\alpha, \theta)\) or \(T_{g}^{2}(\alpha, \theta)\) is used based on the backbone type. For satellite images, we use the discrete rotation \(T_{s}^{3}(p)\) in main experiments. 
Additionally, for the curriculum scheduling function $f(x)$, we adopt the linear variant $f_1(x)$. Ablations of the three satellite rotation variants \(T_s^1(\phi)\), \(T_s^2(\phi)\), \(T_s^3(p)\) and exponential scheduling function \(f_2(x)\), \(f_3(x)\) are provided in the \textit{supplementary material}.
The $\theta_{\text{init}}$, $\theta_{\text{final}}$ are set to 360\textdegree, 70\textdegree, and the $p_{\text{init}}$, $p_{\text{final}}$ are set to 0\% and 75\%.  We choose Sample4Geo \cite{sample4geo} as our baseline and train the model for 80 epochs with a batch size of 16. AdamW optimizer \cite{adam} is applied with an initial learning rate of 0.0001 and a cosine learning rate scheduler. 

\subsection{Comparison with State-of-the-art}

\paragraph{CVUSA\&CVACT:} We evaluate SinGeo under scenarios involving both unknown orientations and limited FoVs on the CVUSA and CVACT validation sets in Tab.~\ref{tab:fov_comparison}. Without FoV-specific training, our single model achieves  SOTA performance across all scenarios on both CVUSA and CVACT. Moreover, SinGeo uses a single CNN backbone that surpasses multiple FoV-specialized trained models of DSM and ConGeo. Although there is a slight drop compared to ConGeo at FoV$=180^\circ$, SinGeo exhibits significant improvements over ConGeo's FoV-specific models in extreme cases of $90^\circ$ and $70^\circ$. Notably, on the CVUSA dataset, SinGeo is the first model to surpass $70\%$ and $50\%$ on R@1 at FoV$=90^\circ$ and FoV$=70^\circ$. We attribute this to the knowledge acquired during the early stages of curriculum learning, which substantially facilitates learning in extreme FoV conditions. Results on the CVACT test set are provided in the \textit{supplementary material}.

\begin{table}[t]
	\centering
	\setlength{\tabcolsep}{1.5pt}  
	\caption{Comparison of state-of-the-art methods on CVUSA and CVACT datasets with north-aligned and center-aligned panorama-satellite pairs. The second-best performance is underlined.}
	\begin{tabular}{@{}l|cc|cc|cc@{}}
		\toprule
		\multirow{2}{*}{Methods} & \multicolumn{2}{c|}{CVUSA} & \multicolumn{2}{c|}{CVACT Val} & \multicolumn{2}{c}{CVACT Test} \\
		& R@1 & R@1\% & R@1 & R@1\% & R@1 & R@1\% \\
		\hline
		DSM \cite{dsm}         & 92.0    & 99.7    & 82.5    & 97.3    & -       & -       \\
		DeHi \cite{dehi}        & 94.3   & \underline{99.8}    & 85.0    & 98.6    & -       & -       \\
		TransGeo \cite{transgeo}    & 94.1    & \underline{99.8}    & 85.0    & 98.4    & -       & -       \\
		GeoDTR \cite{geodtr}     & 95.4    & \textbf{99.9} & 86.2    & \underline{98.8} & 64.5    & \textbf{98.7} \\
		SAIG-D \cite{saig}     & 96.3    & \textbf{99.9} & \underline{89.1}    & \textbf{98.9} & 67.5    & 96.8    \\
		Sample4Geo \cite{sample4geo} & \textbf{98.7} & \textbf{99.9} & \textbf{90.8} & \underline{98.8} & \textbf{71.5} & \textbf{98.7} \\
		\textbf{SinGeo}      & \underline{97.3} & \textbf{99.9} & 87.1 & 96.9 & \underline{69.6} & \underline{97.1} \\
		\bottomrule
	\end{tabular}
	\label{tab:northalign}
\end{table}

Besides, we conduct experiments shown in Tab.~\ref{tab:northalign} to demonstrate that SinGeo achieves superior robust CVGL without compromising its capability for traditional CVGL, where SinGeo still achieves competitive retrieval ability. The slight performance gap with Sample4Geo may be attributed to the latter's specialized sampling strategy.

\begin{table}[t]
	\centering
	\small
	\setlength{\tabcolsep}{0.2pt}
	\caption{Comparison of methods on cross-area and same-area splits of VIGOR under unknown orientation and Limited FoV settings.}
	\label{tab:vigor}
	\begin{tabular}{@{}l|cc|cc|cc|cc@{}}
		\toprule
		\multirow{3}{*}{Methods} & \multicolumn{4}{c|}{\textbf{Cross-Area}} & \multicolumn{4}{c}{\textbf{Same-Area}} \\
		& \multicolumn{2}{c|}{FoV$=360^\circ$} & \multicolumn{2}{c|}{FoV$=90^\circ$} & \multicolumn{2}{c|}{FoV$=360^\circ$} & \multicolumn{2}{c}{FoV$=90^\circ$} \\
		& R@1 & R@1\% & R@1 & R@1\% & R@1 & R@1\% & R@1 & R@1\% \\
		\hline
		VIGOR \cite{vigor}    & 1.4  & 44.6 & -    & -    & 19.1 & 95.1 & -    & -    \\
		TransGeo \cite{transgeo} & 5.5  & 66.9 & -    & -    & 47.7 & \textbf{99.3} & -    & -    \\
		Sample4Geo \cite{sample4geo} & 9.0  & 43.7 & 0.5  & 21.6 & 14.2 & 54.9 & 1.1  & 30.6 \\
		ConGeo \cite{congeo}        & 16.2 & 72.9 & 3.9  & 54.3 & 61.9 & 98.4 & 8.5  & 68.7 \\
		\textbf{SinGeo}         & \textbf{24.7} & \textbf{85.7} & \textbf{4.9} & \textbf{63.7} & \textbf{62.9} & 98.4 & \textbf{24.0} & \textbf{91.5} \\
		\bottomrule
	\end{tabular}
\end{table}

\paragraph{VIGOR.}  Previous research on robust CVGL has predominantly focused on evaluating models using center-aligned datasets, with limited exploration of non-center-aligned datasets such as VIGOR. Tab.~\ref{tab:vigor} highlights the effectiveness of our approach on VIGOR. In the Same-Area subset, SinGeo outperforms existing methods with a remarkable 24.0\% of R@1 and 91.5\% of R@1\% at $90^\circ$. This result once again demonstrates SinGeo's robustness under extreme FoV conditions. In the Cross-Area subset, SinGeo shows comprehensive improvements compared with other methods, which reflects SinGeo's  ability to generalize across geographic regions under challenging conditions. Results under other FoVs are provided in the \textit{supplementary material}.


\paragraph{University-1652.}  Without panorama provided, we use two distinct ground-view images captured at the same location, following \cite{congeo}, denoted as \((I_g^1, I_g^2)\), to replace \((I_g, I_g^*)\) in our learning objective, and only activate the satellite image rotation transformation during training. As shown in Tab.~\ref{tab:university}, due to high task difficulty and limited data, LPN and the University-1652 baseline perform worse than SinGeo and ConGeo. Through a richer contrastive learning pipeline, SinGeo enables more effective feature extraction and achieves better overall results under data scarcity and constraints.


\begin{table}[t]
	\centering
	\small
	\setlength{\tabcolsep}{1pt}  
	\caption{Comparison of methods on St2S and S2St tasks of University-1652 dataset.}
	\begin{tabular}{@{}l|cc|cc@{}}
		\toprule
		\multirow{2}{*}{Methods} & \multicolumn{2}{c|}{{St2S}} & \multicolumn{2}{c}{{S2St}} \\
		& R@1 & R@10 & R@1 & R@10 \\
		\hline
		University-1652 \cite{university} & 0.6  & 5.51 & 0.9  & 6.0  \\
		LPN \cite{lpn}             & 0.7  & -    & 1.4  & -    \\
		ConGeo \cite{congeo}          & \textbf{5.9}  & -    & 6.8  & -    \\
		\textbf{SinGeo}  & \textbf{5.9}  & \textbf{23.73} & \textbf{7.1}    & \textbf{22.8} \\
		\bottomrule
	\end{tabular}
	\label{tab:university}
\end{table}

\begin{figure}[t]
	\centering
	\includegraphics[width=\linewidth]{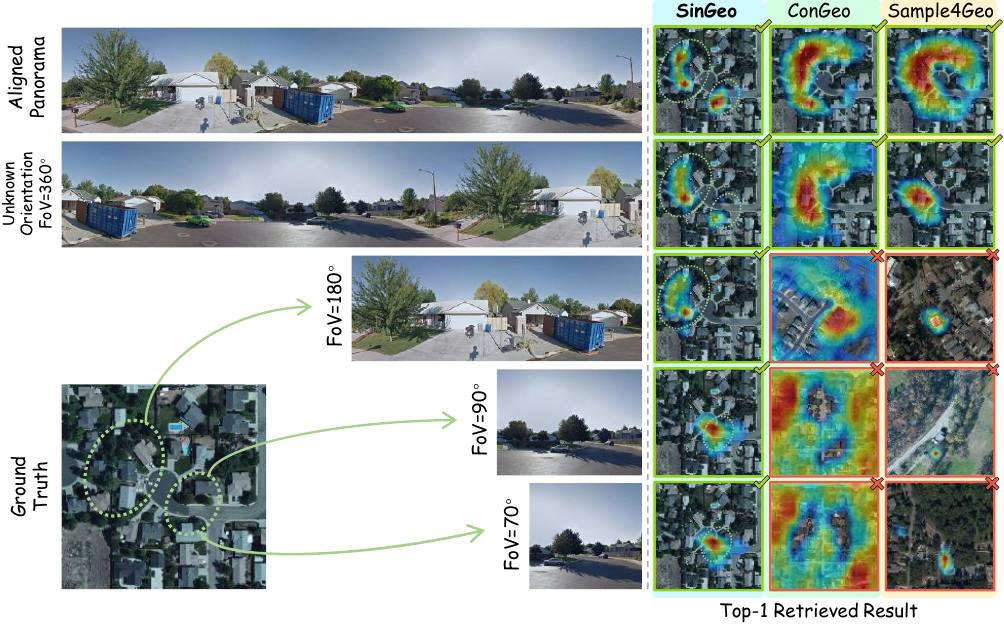}
	\caption{Qualitative evaluation on a sample of CVUSA dataset. Left: A query image under varied orientations and FoVs, together with its ground-truth satellite image. Right: Top-1 retrieved images and activated heatmap regions of different methods. Green circles denote the regions on the satellite image that correspond to the limited-FoV ground images.}
	\label{fig:consis-eval}
\end{figure}

\subsection{Consistency Evaluation}

We conduct consistency evaluation starting from a single sample. Fig.~\ref{fig:consis-eval} shows that SinGeo achieves the best performance with consistent heatmap activations. For panoramas with unknown orientations, although other methods retrieve correctly, their focused regions shift or diverge, which is undesirable since the two panoramas contain the same information. When the FoV narrows, only SinGeo retrieves correct results because its focused regions remain well aligned with the corresponding ground areas. Furthermore, the regions that SinGeo focuses on for limited-FoV images are also focused for panoramas, showing that SinGeo can extend its capability from north-aligned panoramas to other scenarios.

Besides, we conduct the consistency quantitative evaluation on CVUSA validation dataset of 8884 image pairs in Tab.~\ref{tab:consistency}. SinGeo maintains the highest consistency in both satellite and ground branches under different variations.

We argue that the high consistency of SinGeo is highly correlated with its strong recall performance under unknown orientations and limited FoVs, as SinGeo can extend its excellent performance in north-aligned scenarios to other situations. This finding is intended to inspire future research enabling models to maintain higher consistency through novel designs, thereby achieving better robustness.

\begin{table}[t]
	\small
	\centering
	\setlength{\tabcolsep}{5pt}
	\caption{Quantitative evaluation of orientation-consistency(OC) and FoV-consistency(FC) on CVUSA dataset.}
	\begin{tabular}{lcccc}
		\toprule
		{Method} & {$\text{OC}_\text{grd}$} & {$\text{OC}_\text{sat}$} & {$\text{FC}_\text{grd}$} & {$\text{FC}_\text{sat}$} \\
		\midrule
		Sample4Geo \cite{sample4geo}             & 0.65 & 0.67 & 0.50 & 0.66 \\
		ConGeo \cite{congeo}         & 0.38 & 0.71 & 0.35 & 0.68 \\
		\textbf{SinGeo}      & \textbf{0.81} & \textbf{0.92} & \textbf{0.66} & \textbf{0.76} \\
		\bottomrule
	\end{tabular}
	\label{tab:consistency}
\end{table}

\subsection{Cross-architecture Generalization}

The strategies of SinGeo also supports generalization. We transfer the training strategy of SinGeo into different architectures to boost their robustness. In particular, we selected two backbone variants of Sample4Geo---ViT \cite{vit} and ConvNeXt \cite{sample4geo}, and a CNN-Attention hybrid model, GeoDTR \cite{geodtr}. Besides, we compared the boosting efficacy with another plug-and-play method, ConGeo \cite{congeo}. Since ConGeo only supports FoV-fixed training, following its original paper, we trained ConGeo at FoV=180°, enabling it to achieve the highest average R@1.
\begin{table}[t]
	\centering
	\small
	\setlength{\tabcolsep}{1pt}
	\caption{Cross-architecture generalization on the CVUSA dataset under unknown orientation and limited FoV settings.}
	\begin{tabular}{l|cc|cc|cc}
		\toprule
		\multirow{2}{*}{Methods} & \multicolumn{2}{c|}{FoV$=360^\circ$} & \multicolumn{2}{c|}{FoV$=180^\circ$} & \multicolumn{2}{c}{FoV$=90^\circ$} \\
		& R@1 & R@1\% & R@1 & R@1\% & R@1 & R@1\% \\
		\midrule
		GeoDTR \cite{geodtr}       & 7.1  & 30.6 & 3.1  & 19.6 & 0.5  & 8.9  \\
		+ConGeo \cite{congeo}      & 44.5 & 96.2 & 32.9 & 92.9 & 13.3 & 71.4 \\
		\textbf{+SinGeo}      & \textbf{54.8} & \textbf{98.2} & \textbf{37.1} & \textbf{95.9} & \textbf{17.8} & \textbf{82.8} \\
		\midrule
		Sample4Geo[CNN] \cite{sample4geo} & 93.3 & 99.1 & 84.6 & 99.5 & 55.1 & 96.6 \\
		+ConGeo \cite{congeo} & 85.2 & 98.9 & \textbf{92.3} & \textbf{99.7} & 55.9 & 90.9 \\
		\textbf{+SinGeo} & \textbf{96.8} & \textbf{99.5} & {91.8} & {99.5} & \textbf{70.1} & \textbf{97.5} \\
		\midrule
		Sample4Geo[ViT] \cite{sample4geo} & 16.7 & 45.6 & 11.3 & 41.7 & 3.7 & 29.6 \\
		+ConGeo \cite{congeo} & 19.7 & 68.1 & 65.8 & 94.9 & 34.5 & 80.2 \\
		\textbf{+SinGeo} & \textbf{76.0} & \textbf{96.5} & \textbf{72.4} & \textbf{97.2} & \textbf{44.8} & \textbf{89.6} \\
		\bottomrule
	\end{tabular}

	\label{tab:cross-arch}
\end{table}

In Tab.~\ref{tab:cross-arch}, SinGeo exhibits better robustness across architectures than ConGeo and baselines, with significant enhancements on ViT variant of Sample4Geo, boosting R@1 from 16.7\% to 76.0\% at 360\textdegree FoV. Also notably, SinGeo elevates GeoDTR’s R@1 by more than 47\% under 360\textdegree FoV. These results validate the advantage of the curriculum-inspired FoV training paradigm over fixed-FoV paradigms.

\subsection{Ablation Study}

To validate the efficacy of the proposed dual discriminative learning (DDL) architecture and curriculum learning (CL) strategy, we perform an ablation study on the CVUSA dataset to verify their contributions to robustness. As shown in Tab.~\ref{tab:ablation}, incorporating $ I_g^* $ alone enhances performance at 180\textdegree FoV with 92.3\% of R@1, but degrades at 360\textdegree and 90\textdegree compared to the baseline. Adding both $ I_g^* $ and $ I_s^* $  improves 360\textdegree R@1\% but harms performance under limited FoVs. CL boosts robustness, particularly at 90° , and yields the best overall results combined with full DDL. The ablation confirms the synergy between DDL and CL for consistent cross-FoV performance. Ablations among different rotation transformations, scheduling functions and the weights $\omega_1$, $\omega_2$, $\omega_3$, $\gamma$ are provided in the \textit{supplementary material}.

\begin{table}[t]
	\centering
	\small
	\setlength{\tabcolsep}{4pt}
	\caption{Ablations under different DDL and CL configurations on the CVUSA dataset.}
	\begin{tabular}{cc c| *{2}{c}| *{2}{c}| *{2}{c}}  
		\toprule
		\multicolumn{2}{c}{DDL} & \multirow{2}{*}{CL} & \multicolumn{2}{c|}{FoV$=360^\circ$} & \multicolumn{2}{c|}{FoV$=180^\circ$} & \multicolumn{2}{c}{FoV$=90^\circ$} \\
		\( I_g^* \) & \( I_s^* \) &  & R@1 & R@1\% & R@1 & R@1\% & R@1 & R@1\% \\
		\hline
		$\times$ & $\times$ & $\times$ & 93.3 & 99.1 & 84.6 & 99.5 & 55.1 & 96.6 \\
		$\checkmark$ & $\times$ & $\times$ & 85.2 & 98.9 & \textbf{92.3} & \textbf{99.7} & 55.9 & 90.9 \\
		$\checkmark$ & $\checkmark$ & $\times$ & 91.5 & 99.3 & 80.6 & 98.3 & 47.8 & 87.7 \\
		$\checkmark$ & $\times$ & $\checkmark$ & 96.2 & \textbf{99.5} & 92.1 & 99.5 & 66.9 & 96.4 \\
		$\checkmark$ & $\checkmark$ & $\checkmark$ & \textbf{96.8} & \textbf{99.5} & 91.8 & 99.5 & \textbf{70.1} & \textbf{97.5} \\
		\bottomrule
	\end{tabular}
	\label{tab:ablation}
\end{table}

\section{Conclusion and Limitation}

In this work, we presented SinGeo, a new paradigm for robust CVGL. SinGeo leverages a dual discriminative learning architecture, treating both ground and satellite branches equally and strengthening intra-view discriminability. SinGeo also introduces a curriculum learning strategy guiding the model in a natural manner. Extensive experiments on four benchmarks indicate that SinGeo uses only a single backbone and surpasses existing methods by a large margin under varying orientations and FoVs. These results provide a clear answer to the question raised in our introduction: a single backbone can indeed achieve consistent and high performance across diverse conditions, without additional modules or explicit transformations. Moreover, SinGeo can be transferred to other CVGL methods to further boost their robustness, outperforming FoV-specialized paradigm enhancement. 
Beyond performance, our consistency evaluation provides a perspective for understanding robustness in CVGL, showing that stable model responses across scenarios are highly correlated with robust retrieval.

Admittedly, SinGeo requires the prior knowledge of panoramas during training. How to achieve excellent performance using datasets without aligned panoramas, such as University-1652 \cite{university}, remains a challenge in CVGL.

{
    
    \bibliographystyle{ieeenat_fullname}
    \bibliography{main}

@String(CVPR= {IEEE Conf. Comput. Vis. Pattern Recog.})

@String(AAAI = {AAAI})

@String(CVM = {Computational Visual Media})

@String(CVPR  = {CVPR})

@inproceedings{cvm,
	title={Cvm-net: Cross-view matching network for image-based ground-to-aerial geo-localization},
	author={Hu, Sixing and Feng, Mengdan and Nguyen, Rang MH and Lee, Gim Hee},
	booktitle={Proceedings of the IEEE Conference on Computer Vision and Pattern Recognition},
	pages={7258--7267},
	year={2018}
}

@inproceedings{cvft,
	title={Optimal feature transport for cross-view image geo-localization},
	author={Shi, Yujiao and Yu, Xin and Liu, Liu and Zhang, Tong and Li, Hongdong},
	booktitle={Proceedings of the AAAI Conference on Artificial Intelligence},
	volume={34},
	number={07},
	pages={11990--11997},
	year={2020}
}

@inproceedings{lin2013cross,
	title={Cross-view image geolocalization},
	author={Lin, Tsung-Yi and Belongie, Serge and Hays, James},
	booktitle={Proceedings of the IEEE Conference on Computer Vision and Pattern Recognition},
	pages={891--898},
	year={2013}
}

@inproceedings{li2019navi,
	title={Cross-view policy learning for street navigation},
	author={Li, Ang and Hu, Huiyi and Mirowski, Piotr and Farajtabar, Mehrdad},
	booktitle={Proceedings of the IEEE/CVF international conference on computer vision},
	pages={8100--8109},
	year={2019}
}

@article{dehi,
	title={DeHi: A decoupled hierarchical architecture for unaligned ground-to-aerial geo-localization},
	author={Wang, Teng and Li, Jiawen and Sun, Changyin},
	journal={IEEE Transactions on Circuits and Systems for Video Technology},
	volume={34},
	number={3},
	pages={1927--1940},
	year={2023},
	publisher={IEEE}
}

@inproceedings{transgeo,
	title={Transgeo: Transformer is all you need for cross-view image geo-localization},
	author={Zhu, Sijie and Shah, Mubarak and Chen, Chen},
	booktitle={Proceedings of the IEEE/CVF Conference on Computer Vision and Pattern Recognition},
	pages={1162--1171},
	year={2022}
}

@inproceedings{sample4geo,
	title={Sample4geo: Hard negative sampling for cross-view geo-localisation},
	author={Deuser, Fabian and Habel, Konrad and Oswald, Norbert},
	booktitle={Proceedings of the IEEE/CVF International Conference on Computer Vision},
	pages={16847--16856},
	year={2023}
}

@inproceedings{geodtr,
	title={Cross-view geo-localization via learning disentangled geometric layout correspondence},
	author={Zhang, Xiaohan and Li, Xingyu and Sultani, Waqas and Zhou, Yi and Wshah, Safwan},
	booktitle={Proceedings of the AAAI conference on artificial intelligence},
	volume={37},
	number={3},
	pages={3480--3488},
	year={2023}
}

@inproceedings{cvusa,
	title={Predicting ground-level scene layout from aerial imagery},
	author={Zhai, Menghua and Bessinger, Zachary and Workman, Scott and Jacobs, Nathan},
	booktitle={Proceedings of the IEEE Conference on Computer Vision and Pattern Recognition},
	pages={867--875},
	year={2017}
}

@inproceedings{cvact,
	title={Lending orientation to neural networks for cross-view geo-localization},
	author={Liu, Liu and Li, Hongdong},
	booktitle={Proceedings of the IEEE/CVF Conference on Computer Vision and Pattern Recognition},
	pages={5624--5633},
	year={2019}
}

@inproceedings{congeo,
	title={ConGeo: Robust Cross-view Geo-localization across Ground View Variations},
	author={Mi, Li and Xu, Chang and Castillo-Navarro, Javiera and Montariol, Syrielle and Yang, Wen and Bosselut, Antoine and Tuia, Devis},   booktitle={European Conference on Computer Vision},
	year={2024}
}

@inproceedings{dsm,
	title={Where am i looking at? joint location and orientation estimation by cross-view matching},
	author={Shi, Yujiao and Yu, Xin and Campbell, Dylan and Li, Hongdong},
	booktitle={Proceedings of the IEEE/CVF Conference on Computer Vision and Pattern Recognition},
	pages={4064--4072},
	year={2020}
}

@inproceedings{arcgeo,
	title={Arcgeo: Localizing limited field-of-view images using cross-view matching},
	author={Shugaev, Maxim and Semenov, Ilya and Ashley, Kyle and Klaczynski, Michael and Cuntoor, Naresh and Lee, Mun Wai and Jacobs, Nathan},
	booktitle={Proceedings of the IEEE/CVF Winter Conference on Applications of Computer Vision},
	pages={209--218},
	year={2024}
}

@inproceedings{gal,
	title={Global assists local: Effective aerial representations for field of view constrained image geo-localization},
	author={Rodrigues, Royston and Tani, Masahiro},
	booktitle={Proceedings of the IEEE/CVF Winter Conference on Applications of Computer Vision},
	pages={3871--3879},
	year={2022}
}

@inproceedings{bevcv,
	title={BEV-CV: Birds-eye-view transform for cross-view geo-localisation},
	author={Shore, Tavis and Hadfield, Simon and Mendez, Oscar},
	booktitle={2024 IEEE/RSJ International Conference on Intelligent Robots and Systems (IROS)},
	pages={11048--11055},
	year={2024},
	organization={IEEE}
}

@inproceedings{cl,
	title={Curriculum learning},
	author={Bengio, Yoshua and Louradour, J{\'e}r{\^o}me and Collobert, Ronan and Weston, Jason},
	booktitle={Proceedings of the 26th annual international conference on machine learning},
	pages={41--48},
	year={2009}
}

@inproceedings{
	vit,
	title={An Image is Worth 16x16 Words: Transformers for Image Recognition at Scale},
	author={Alexey Dosovitskiy and Lucas Beyer and Alexander Kolesnikov and Dirk Weissenborn and Xiaohua Zhai and Thomas Unterthiner and Mostafa Dehghani and Matthias Minderer and Georg Heigold and Sylvain Gelly and Jakob Uszkoreit and Neil Houlsby},
	booktitle={International Conference on Learning Representations},
	year={2021},
	url={https://openreview.net/forum?id=YicbFdNTTy}
}

@inproceedings{convnext,
	title={A convnet for the 2020s},
	author={Liu, Zhuang and Mao, Hanzi and Wu, Chao-Yuan and Feichtenhofer, Christoph and Darrell, Trevor and Xie, Saining},
	booktitle={Proceedings of the IEEE/CVF Conference on Computer Vision and Pattern Recognition},
	pages={11976--11986},
	year={2022}
}

@inproceedings{softmargin,
	title={Localizing and orienting street views using overhead imagery},
	author={Vo, Nam N and Hays, James},
	booktitle={European conference on computer vision},
	pages={494--509},
	year={2016},
	organization={Springer}
}

@article{safa,
	title={Spatial-aware feature aggregation for cross-view image based geo-localization},
	author={Shi, Yujiao and Liu, Liu and Yu, Xin and Li, Hongdong},
	journal={Advances in Neural Information Processing Systems},
	volume={32},
	year={2019}
}

@inproceedings{epbev,
	title={Cross-view image geo-localization with Panorama-BEV Co-Retrieval Network},
	author={Ye, Junyan and Lv, Zhutao and Li, Weijia and Yu, Jinhua and Yang, Haote and Zhong, Huaping and He, Conghui},
	booktitle={European Conference on Computer Vision},
	pages={74--90},
	year={2024},
	organization={Springer}
}

@article{infonce,
	title={Representation learning with contrastive predictive coding},
	author={Oord, Aaron van den and Li, Yazhe and Vinyals, Oriol},
	journal={arXiv preprint arXiv:1807.03748},
	year={2018}
}

@article{shortcut,
	title={Can contrastive learning avoid shortcut solutions?},
	author={Robinson, Joshua and Sun, Li and Yu, Ke and Batmanghelich, Kayhan and Jegelka, Stefanie and Sra, Suvrit},
	journal={Advances in neural information processing systems},
	volume={34},
	pages={4974--4986},
	year={2021}
}

@inproceedings{vigor,
	title={Vigor: Cross-view image geo-localization beyond one-to-one retrieval},
	author={Zhu, Sijie and Yang, Taojiannan and Chen, Chen},
	booktitle={Proceedings of the IEEE/CVF Conference on Computer Vision and Pattern Recognition},
	pages={3640--3649},
	year={2021}
}

@inproceedings{university,
	title={University-1652: A multi-view multi-source benchmark for drone-based geo-localization},
	author={Zheng, Zhedong and Wei, Yunchao and Yang, Yi},
	booktitle={Proceedings of the 28th ACM international conference on Multimedia},
	pages={1395--1403},
	year={2020}
}

@inproceedings{gradcam,
	title={Grad-cam: Visual explanations from deep networks via gradient-based localization},
	author={Selvaraju, Ramprasaath R and Cogswell, Michael and Das, Abhishek and Vedantam, Ramakrishna and Parikh, Devi and Batra, Dhruv},
	booktitle={Proceedings of the IEEE international conference on computer vision},
	pages={618--626},
	year={2017}
}

@article{ssim,
	title={Image quality assessment: from error visibility to structural similarity},
	author={Wang, Zhou},
	journal={IEEE Transactions on Image Processing},
	volume={13},
	number={4},
	pages={600--612},
	year={2004},
	publisher={IEEE}
}

@article{seh,
	title={Soft exemplar highlighting for cross-view image-based geo-localization},
	author={Guo, Yulan and Choi, Michael and Li, Kunhong and Boussaid, Farid and Bennamoun, Mohammed},
	journal={IEEE transactions on image processing},
	volume={31},
	pages={2094--2105},
	year={2022},
	publisher={IEEE}
}

@article{saig,
	title={Simple, effective and general: A new backbone for cross-view image geo-localization},
	author={Zhu, Yingying and Yang, Hongji and Lu, Yuxin and Huang, Qiang},
	journal={arXiv preprint arXiv:2302.01572},
	year={2023}
}

@article{lpn,
	title={Each part matters: Local patterns facilitate cross-view geo-localization},
	author={Wang, Tingyu and Zheng, Zhedong and Yan, Chenggang and Zhang, Jiyong and Sun, Yaoqi and Zheng, Bolun and Yang, Yi},
	journal={IEEE Transactions on Circuits and Systems for Video Technology},
	volume={32},
	number={2},
	pages={867--879},
	year={2021},
	publisher={IEEE}
}

@inproceedings{clcvgl,
	title={Unleashing unlabeled data: A paradigm for cross-view geo-localization},
	author={Li, Guopeng and Qian, Ming and Xia, Gui-Song},
	booktitle={Proceedings of the IEEE/CVF Conference on Computer Vision and Pattern Recognition},
	pages={16719--16729},
	year={2024}
}

@inproceedings{adam,
	title={Decoupled Weight Decay Regularization},
	author={Ilya Loshchilov and Frank Hutter},
	booktitle={International Conference on Learning Representations},
	year={2017},
	url={https://api.semanticscholar.org/CorpusID:53592270}
}

@incollection{pcc,
	title={Pearson correlation coefficient},
	author={Benesty, Jacob and Chen, Jingdong and Huang, Yiteng and Cohen, Israel},
	booktitle={Noise reduction in speech processing},
	pages={1--4},
	year={2009},
	publisher={Springer}
}

@INPROCEEDINGS{triplet,
	author={Schroff, Florian and Kalenichenko, Dmitry and Philbin, James},
	booktitle={2015 IEEE Conference on Computer Vision and Pattern Recognition (CVPR)}, 
	title={FaceNet: A unified embedding for face recognition and clustering}, 
	year={2015},
	pages={815-823},
	doi={10.1109/CVPR.2015.7298682}}
}


\clearpage
\setcounter{page}{1}
\maketitlesupplementary
\renewcommand{\thesection}{\Alph{section}}
\setcounter{section}{0} 






In this supplementary material, we provide the following information for the completeness of our paper.

\begin{itemize}\item Ablations of different rotation transformations (Section \ref{sec:rot}).\item Ablations of different scheduling functions (Section \ref{sec:schf}).\item Ablations of different loss weights (Section \ref{sec:weights}).\item Performance on CVACT test set (Section \ref{sec:cvacttest}).\item Supplementary results on VIGOR (Section \ref{sec:supvigor}).\item Consistency metrics based on cosine similarity and pearson correlation coefficient (Section \ref{sec:consismetric}).\item Additional qualitative visualization results (Section \ref{sec:additionalvis}).\item Implementation details (Section \ref{sec:imple}).\end{itemize}

\section{Ablations of Different Rotation Transformations}\label{sec:rot} 

\begin{table}[htbp]
	\centering
	\small
	\setlength{\tabcolsep}{2pt}
	\caption{Performance comparison of different rotation transformations for $ I_s $ on the CVUSA dataset.}
	\begin{tabular}{c|*{2}{c}|*{2}{c}|*{2}{c}}  
		\toprule
		\multirow{2}{*}{Transformations} & \multicolumn{2}{c|}{FoV$=360^\circ$} & \multicolumn{2}{c|}{FoV$=180^\circ$} & \multicolumn{2}{c}{FoV$=90^\circ$} \\
		& R@1 & R@1\% & R@1 & R@1\% & R@1 & R@1\% \\
		\hline
		$ T_s^1(\phi) $ & 89.2 & 97.3 & 77.5 & 94.2 & 63.7 & 96.4 \\
		$ T_s^2(\phi) $ & 88.9 & 97.5 & 81.1 & 97.4 & 60.7 & 95.8 \\
		$ T_s^3(p) $    & \textbf{96.8} & \textbf{99.5} & \textbf{91.8} & \textbf{99.5} & \textbf{70.1} & \textbf{97.5} \\
		\bottomrule
	\end{tabular}
	\label{tab:transformations}
\end{table}
We compare the performance of proposed \( T_s^1(\phi) \), \( T_s^2(\phi) \), and \( T_s^3(p) \) on CVUSA dataset. In Tab.~\ref{tab:transformations}, the discrete rotation transformation \( T_s^3(p) \) consistently outperforms the continuous variants \( T_s^1(\phi) \) and \( T_s^2(\phi) \) across all FoVs on the CVUSA dataset. We assume that discrete rotation transformations outperform continuous ones because the former can avoid introducing additional padding boundaries and will not cause information loss. Discrete 90° multiples enable exact pixel permutations, preserving discriminative features.

\section{Ablations of Different Scheduling Functions}\label{sec:schf}
\begin{table}[htbp]
	\centering
	\small
	\setlength{\tabcolsep}{2pt}
	\caption{Performance comparison of different scheduling functions on the CVUSA dataset.}
	\begin{tabular}{c|c|cc|cc|cc} 
		\toprule
		\multirow{2}{*}{\makecell{Scheduling\\Function}} & \multirow{1}{*}{Avg.} & \multicolumn{2}{c|}{FoV$=360^\circ$} & \multicolumn{2}{c|}{FoV$=180^\circ$} & \multicolumn{2}{c}{FoV$=90^\circ$} \\
		& R@1 & R@1 & R@1\% & R@1 & R@1\% & R@1 & R@1\% \\
		\hline
		$f_1(x)$ & \textbf{86.2} & 96.8 & 99.5 & \textbf{91.8} & \textbf{99.5} & \textbf{70.1} & \textbf{97.5} \\
		$f_2(x)(\lambda=3)$ & 79.9 & 94.6 & 99.6 & 83.7 & 99.2 & 61.5 & 96.0 \\
		$f_3(x)(\lambda=3)$ & 82.8 & \textbf{97.1} & \textbf{99.8} & 90.9 & 99.3 & 60.4 & 92.3 \\
		$f_2(x)(\lambda=5)$ & 82.0 & 95.8 & 99.5 & 83.1 & 98.3 & 67.0 & 96.2 \\
		$f_3(x)(\lambda=5)$ & 82.3 & 96.6 & 99.8 & 89.8 & 99.6 & 60.5 & 94.5 \\
		$\textit{Random}$ & 76.7 & 90.3 & 98.7 & 83.6 & 98.8 & 56.2 & 93.5 \\
		\bottomrule
	\end{tabular}
	\label{tab:scheduling-functions}
\end{table}

We also considered different scheduling functions. As observed in Tab.~\ref{tab:scheduling-functions}, a simple linear scheduling function $f_1(x)$ achieves superior overall performance with 86.2 of average R@1 compared to alternative exponential variants $f_2(x)$, $f_3(x)$ and random scheduling. Notably, all scheduling functions that progress from easy to hard outperform random scheduling, as the latter lacks a progressive adaptation mechanism. We hypothesize that the linear schedule maintains a constant gradient in training difficulty between epochs, thereby mitigating abrupt shifts that could lead to model instability, such as insufficient adaptation to escalating challenges. This result validates that a straightforward progressive training paradigm is highly beneficial for model to learn robustness.

\section{Ablations of Different Loss Weights}\label{sec:weights}

\begin{table}[htbp]
	\centering
	\small
	\setlength{\tabcolsep}{1.5pt}
	\caption{Performance comparison of different loss weight configurations on the CVUSA dataset.}
	\begin{tabular}{cc|c|cc|cc|cc}  
		\toprule
		\multirow{2}{*}{$\gamma$} & \multirow{2}{*}{$\omega_1\omega_2\omega_3$} & \multirow{1}{*}{Avg.} & \multicolumn{2}{c|}{FoV$=360^\circ$} & \multicolumn{2}{c|}{FoV$=180^\circ$} & \multicolumn{2}{c}{FoV$=90^\circ$} \\
		& &R@1 & R@1 & R@1\% & R@1 & R@1\% & R@1 & R@1\% \\
		\hline
		0.5 & 0.25 & \textbf{86.1} & \textbf{96.8} & 99.5 & \textbf{91.5} & 99.4 & 70.1 & {97.5} \\
		0.5 & 0.5 & 84.7 & 96.2 & 99.6 & 90.0 & 99.4 & 68.0 & \textbf{97.7} \\
		0.25 & 0.5 & 85.6 & 96.4 & \textbf{99.7} & 90.7 & \textbf{99.6} & 69.6 & 97.6 \\
		0.25 & 0.25 & 85.8 & 96.7 & \textbf{99.7} & 90.2 & 99.2 & \textbf{70.6} & 97.6 \\
		\bottomrule
	\end{tabular}
	\label{tab:weight_ablations}
\end{table}

In Tab.~\ref{tab:weight_ablations}, we look into the effect of loss weights \(\gamma\) and \(\omega_1, \omega_2, \omega_3\) on performance. Results show that different weight configurations cause minor performance fluctuations across FoV settings. The default setup \(\gamma=0.5\), \(\omega=0.25\) achieves the best average R@1 of 86.1 and outperforms other configurations at FoV\(=360^\circ\) and FoV\(=180^\circ\).

\section{Performance on CVACT Test Set}\label{sec:cvacttest}

\begin{table*}[!htbp]
	
	\centering
	\small
	\setlength{\tabcolsep}{2.0pt}  
	\caption{Comparison with single-model performance of selected methods on CVACT Test dataset under unknown orientation and limited FoV settings.}
	\begin{tabular}{@{}l|c|cccc|cccc|cccc|cccc@{}}
		\toprule  
		\multirow{2}{*}{\centering Methods}
		& \multirow{1}{*}{\centering \makecell{Avg.}}  
		& \multicolumn{4}{c|}{\makecell{FoV $360^\circ$}} 
		& \multicolumn{4}{c|}{\makecell{FoV $180^\circ$}} 
		& \multicolumn{4}{c|}{\makecell{FoV $90^\circ$}} 
		& \multicolumn{4}{c}{\makecell{FoV $70^\circ$}} \\
		& R@1 & R@1 & R@5 & R@10 & R@1\% & R@1 & R@5 & R@10 & R@1\% & R@1 & R@5 & R@10 & R@1\% & R@1 & R@5 & R@10 & R@1\% \\
		\midrule  
		Sample4Geo\cite{sample4geo} & 2.4 & 8.0 & 12.5 & 14.0 & 31.0 & 1.0 & 2.7 & 3.8 & 25.1 & 0.4 & 1.3 & 2.2 & 24.1 & 0.2 & 0.7 & 1.1 & 16.9 \\
		ConGeo[360]\cite{congeo} & 25.0 & 64.3 & 81.5 & 84.7 & 95.6 & 28.6 & 47.8 & 54.9 & 88.4 & 5.2 & 13.0 & 17.9 & 67.8 & 1.8 & 5.6 & 8.4 & 53.3 \\
		ConGeo[180]\cite{congeo} & 25.4 & 37.4 & 55.3 & 61.1 & 89.6 & 45.6 & \textbf{67.1} & \textbf{73.0} & \textbf{94.3} & 13.7 & 28.3 & 35.6 & 81.7 & 5.0 & 13.1 & 18.0 & 68.6 \\
		
		\textbf{SinGeo} & \textbf{35.8} & \textbf{66.7} & \textbf{84.2} & \textbf{86.8} & \textbf{96.0} & \textbf{46.0} & 66.8 & 72.6 & 93.2 & \textbf{19.6} & \textbf{38.8} & \textbf{47.1} & \textbf{89.3} & \textbf{11.0} & \textbf{25.5} & \textbf{33.3} & \textbf{84.2} \\
		
		\bottomrule  
	\end{tabular}
	
	\label{tab:cvact_test_fov_comparison} 
\end{table*}

In the main paper, we reported the results on CVACT validation set. In this section, we also evaluate SinGeo on the CVACT test set under unknown orientation and limited FoV settings as shown in Tab.~\ref{tab:cvact_test_fov_comparison}, comparing it with Sample4Geo \cite{sample4geo} and two FoV-specialized ConGeo variants \cite{congeo}. ConGeo[360] and ConGeo[180] are trained specifically at 360° and 180° FoV respectively, showing strong performance at their trained angles but significant degradation at unseen narrow FoVs.

In contrast, SinGeo uses a single model without FoV-specific training and achieves the best average R@1 of 35.8 across all FoVs. It outperforms both ConGeo variants at every FoV except 180° where ConGeo[180] holds a marginal edge in R@5 and R@10. Notably, SinGeo maintains best performance even at extreme narrow FoVs.

\section{Supplementary Results on VIGOR}\label{sec:supvigor}

\begin{table*}[!htbp]
	\centering
	\small
	\setlength{\tabcolsep}{1.5pt}  
	\caption{Comparison of methods on cross-area and same-area splits of VIGOR under unknown orientation and limited FoV settings. “–” denotes metrics not provided in the original paper.}
	\begin{tabular}{@{}c|l|cccc|cccc|cccc|cccc@{}}
		\toprule
		\multirow{2}{*}{\centering Set} & \multirow{2}{*}{\centering Methods}
		& \multicolumn{4}{c|}{\makecell{FoV $360^\circ$}} 
		& \multicolumn{4}{c|}{\makecell{FoV $180^\circ$}} 
		& \multicolumn{4}{c|}{\makecell{FoV $90^\circ$}} 
		& \multicolumn{4}{c}{\makecell{FoV $70^\circ$}} \\
		& & R@1 & R@5 & R@10 & R@1\% & R@1 & R@5 & R@10 & R@1\% & R@1 & R@5 & R@10 & R@1\% & R@1 & R@5 & R@10 & R@1\% \\
		\midrule
		\multirow{5}{*}{\rotatebox{90}{Cross-Area}}
		& VIGOR \cite{vigor}    & 1.4 & - & - & 44.6 & - & - & - & - & - & - & - & - & - & - & - & - \\
		& TransGeo \cite{transgeo} & 5.5 & - & - & 66.9 & - & - & - & - & - & - & - & - & - & - & - & - \\
		& Sample4Geo \cite{sample4geo} & 9.0 & - & - & 43.7 & - & - & - & - & 0.5 & - & - & 21.6 & - & - & - & - \\
		& ConGeo \cite{congeo}        & 16.2 & - & - & 72.9 & - & - & - & - & 3.9 & - & - & 54.3 & - & - & - & - \\
		& \textbf{SinGeo}         & \textbf{24.7} & \textbf{43.5} & \textbf{51.4} & \textbf{85.7} & \textbf{11.0} & \textbf{24.0} & \textbf{30.7} & \textbf{72.6} & \textbf{4.9} & \textbf{12.9} & \textbf{18.2} & \textbf{63.7} & \textbf{3.3} & \textbf{9.2} & \textbf{13.2} & \textbf{56.0} \\
		\midrule
		\multirow{5}{*}{\rotatebox{90}{Same-Area}}
		& VIGOR \cite{vigor}    & 19.1 & - & - & 95.1 & - & - & - & - & - & - & - & - & - & - & - & - \\
		& TransGeo \cite{transgeo} & 47.7 & - & - & \textbf{99.3} & - & - & - & - & - & - & - & - & - & - & - & - \\
		& Sample4Geo \cite{sample4geo} & 14.2 & - & - & 54.9 & - & - & - & - & 1.1 & - & - & 30.6 & - & - & - & - \\
		& ConGeo \cite{congeo}        & 61.9 & - & - & 98.4 & - & - & - & - & 8.5 & - & - & 68.7 & - & - & - & - \\
		& \textbf{SinGeo}         & \textbf{62.9} & \textbf{88.5} & \textbf{91.5} & {98.0} & \textbf{43.0} & \textbf{69.0} & \textbf{75.2} & \textbf{94.0} & \textbf{24.0} & \textbf{46.3} & \textbf{54.6} & \textbf{91.5} & \textbf{16.1} & \textbf{34.2} & \textbf{42.3} & \textbf{85.5} \\
		\bottomrule
	\end{tabular}
	\label{tab:vigor_fov_comparison}
\end{table*}

In the main paper, we reported results of VIGOR dataset at 360° and 90° FoV. We further supplement results at 180° and 70° FoV as shown in Tab.~\ref{tab:vigor_fov_comparison}. SinGeo achieves excellent performance on VIGOR, a non-center-aligned dataset, outperforming other compared methods across both Cross-Area and Same-Area splits.

\section{The Consistency Metrics Based on Cosine Similiarity and PCC}\label{sec:consismetric}

\begin{table}[htbp]
	\small
	\centering
	\setlength{\tabcolsep}{2pt}
	\caption{Quantitative evaluation of orientation-consistency(OC) and FoV-consistency(FC) on CVUSA dataset based on cosine similarity and pearson correlation coefficient).}
	\begin{tabular}{l l c c c c}
		\toprule
		{Metric} & {Method} & {$\text{OC}_\text{grd}$} & {$\text{OC}_\text{sat}$} & {$\text{FC}_\text{grd}$} & {$\text{FC}_\text{sat}$} \\
		\midrule
		\multirow{3}{*}{Cosine Similarity} 
		& Sample4Geo \cite{sample4geo} & 0.37 & 0.40 & 0.25 & 0.32 \\
		& ConGeo \cite{congeo}         & 0.15 & 0.61 & 0.22 & 0.46 \\
		& \textbf{SinGeo}              & \textbf{0.77} & \textbf{0.91} & \textbf{0.54} & \textbf{0.63} \\
		\midrule
		\multirow{3}{*}{PCC} 
		& Sample4Geo \cite{sample4geo} & 0.13 & 0.18 & 0.04 & 0.11 \\
		& ConGeo \cite{congeo}         & 0.10 & 0.46 & 0.17 & 0.21 \\
		& \textbf{SinGeo}              & \textbf{0.65} & \textbf{0.89} & \textbf{0.31} & \textbf{0.54} \\
		\bottomrule
	\end{tabular}
	\label{tab:consistency_combined}
\end{table}

In the main paper, we adopt the Structural Similarity Index (SSIM) \cite{ssim}, a classic 2D similarity measurement method, to calculate the consistency between activation heatmaps. To further validate our conclusions, we flatten the 2D heatmaps into 1D tensors and utilize cosine similarity and Pearson Correlation Coefficient (PCC) \cite{pcc} to compute the consistency metrics between these tensors. As shown in Tab.~\ref{tab:consistency_combined}, SinGeo still achieves the highest scores across all orientation-consistency (OC) and FoV-consistency (FC) metrics when using cosine similarity and PCC. These results confirm that SinGeo preserves stable activation under orientation shifts and FoV variations.

\section{Additional Qualitative Visualization Results}\label{sec:additionalvis}

\begin{figure*}[htbp]
	\centering
	\includegraphics[width=0.8\linewidth, page=1]{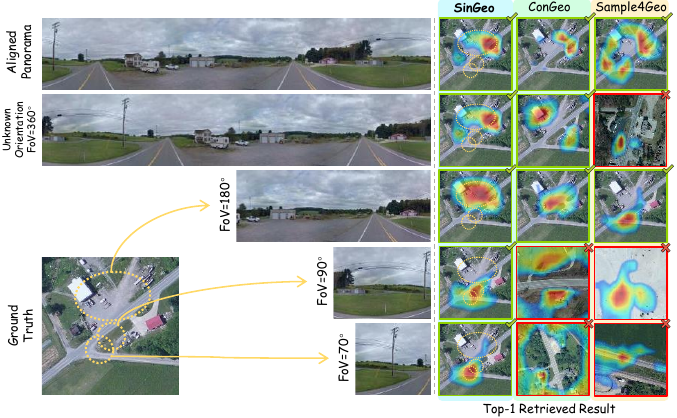}
	\caption{Qualitative visualization result on CVUSA. Yellow circles denote the regions on the satellite image that correspond to the limited-FoV ground images.}
	\label{fig:addvisual1-page1}
\end{figure*}

\begin{figure*}[htbp]
	\centering
	\includegraphics[width=0.8\linewidth, page=2]{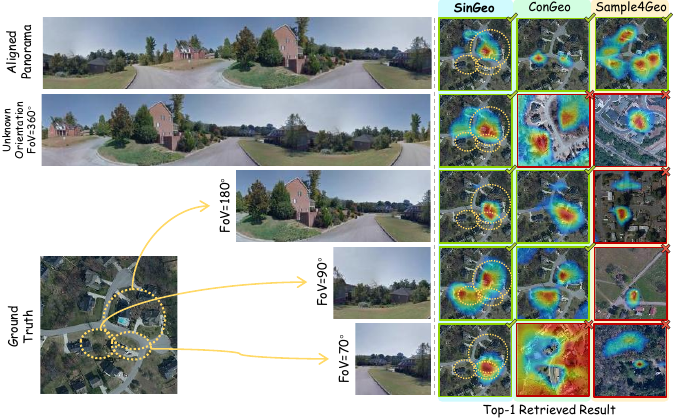}
	\caption{Qualitative visualization result on CVUSA. Yellow circles denote the regions on the satellite image that correspond to the limited-FoV ground images.}
	\label{fig:addvisual1-page2}
\end{figure*}

\begin{figure*}[htbp]
	\centering
	\includegraphics[width=0.8\linewidth, page=3]{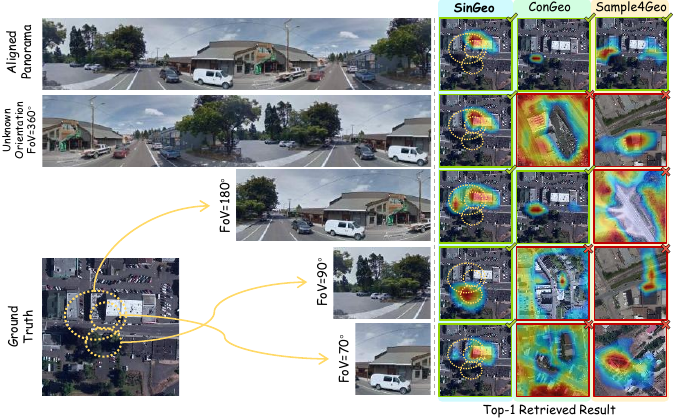}
	\caption{Qualitative visualization result on CVUSA. Yellow circles denote the regions on the satellite image that correspond to the limited-FoV ground images.}
	\label{fig:addvisual1-page3}
\end{figure*}

We further provide supplementary qualitative visualizations in Fig.~\ref{fig:addvisual1-page1}, Fig.~\ref{fig:addvisual1-page2} and Fig.~\ref{fig:addvisual1-page3}. When the aligned panorama is applied with a random translation (the second row), SinGeo maintains consistent responses on satellite images. As the FoV further narrows (last three rows), the model consistently responds to key semantic regions of satellite images that correspond to the ground images (the yellow dashed circles). Compared with SinGeo, the results of ConGeo and Sample4Geo are less satisfactory. Even when retrieval is correct, their response regions show certain shifts or errors. As shown in the first and second rows of Fig.~\ref{fig:addvisual1-page1}, after panorama translation, ConGeo’s responses on satellite images become completely different despite identical information in panoramas. As shown in the third row of Fig.~\ref{fig:addvisual1-page1}, when FoV=180$\circ$, Sample4Geo’s response region does not match the corresponding area of the ground query image even with correct retrieval.

\section{Implementation Details}\label{sec:imple}

We follow the data preprocessing method of our baseline Sample4Geo \cite{sample4geo}. Our code of this paper uses Python 3.8.20, PyTorch 2.1.1, timm 0.9.0, and scikit-learn 1.3.2. All training and testing are conducted on a single NVIDIA A100 GPU.
	
\end{document}